# Convolutional LSTM Neural Networks for Modeling Wildland Fire Dynamics

Running Head: ConvLSTMs for Wildland Fire Dynamics

John Burge[a,c], Matthew Bonanni[b], Matthias Ihme[a,b], and R. Lily Hu[a]

[a]Google Research, Mountain View, CA 94043, USA
[b]Department of Mechanical Engineering, Stanford University, Stanford, CA 94305, USA
[c]Corresponding Author. Email: lawnguy@google.com

## Abstract

As the climate changes, the severity of wildland fires is expected to worsen. Models that accurately capture fire propagation dynamics greatly help efforts for understanding, responding to and mitigating the damages caused by these fires. Machine learning techniques provide a potential approach for developing such models. The objective of this study is to evaluate the feasibility of using a Convolutional Long Short-Term Memory (ConvLSTM) recurrent neural network to model the dynamics of wildland fire propagation. The machine learning model is trained on simulated wildfire data generated by a mathematical analogue model. Three simulated datasets are analyzed, each with increasing degrees of complexity. The simplest dataset includes a constant wind direction as a single confounding factor, whereas the most complex dataset includes dynamic wind, complex terrain, spatially varying moisture content and heterogenous vegetation density distributions. We examine how effective the ConvLSTM can learn the fire-spread dynamics over consecutive time steps. It is shown that ConvLSTMs can capture local fire transmission events, as well as the overall fire dynamics, such as the rate at which the fire spreads. Finally, we demonstrate that ConvLSTMs outperform other network architectures that have previously been used to model similar wildland fire dynamics.

## Summary

In the last decade, remarkable improvements to Deep Neural Networks have been made that have allowed their successful use in many scientific domains, although their application to the geosciences is relatively new. This work demonstrates how such networks can be used to learn wildfire-spread dynamics.

# Contents

# 1 Introduction

Over recent decades, computational models predicting wildfire behavior for fire management, risk assessment, and wildfire mitigation have seen significant advancements (Sullivan 2009a). These models can be categorized as physical, empirical, and mathematical analogues (Sullivan 2009a, b, c). Physical models are built from first principles and provide a high level of fidelity in predicting the fire behavior by solving conservation equations to capture relevant physical processes. Notable examples of these include FIRETEC (Linn and Harlow 1997), IUSTI (Larini *et al.* 1998; Porterie *et al.* 2000), and WFDS (Mell *et al.* 2007). Unfortunately, the computational complexity of these models prevents their application to real-time simulations, so these models are often used in augmenting field experimentation or in the detailed analysis of wildfire dynamics (Sullivan 2009a).

Purely empirical models, in contrast, rely solely on data, making no assumptions about fire behavior based on theory (Sullivan 2009b). Notable examples of this category include FDRS (McArthur 1966, 1967) and CSIRO GSFM (Cheney and Sullivan 1997). In contrast, quasi-empirical models combine observations from real forest fires and laboratory experiments with knowledge of the underlying combustion and heat-transfer processes. The foundational quasi-empirical model of Rothermel (Rothermel 1972) serves as the cornerstone for several fire prediction methods in the United States, including BEHAVE (Andrews 1986).

Finally, simulation and mathematical analogue models attempt to model fire spread over a two-dimensional landscape (Sullivan 2009c). Simulation models, such as FARSITE (Finney 1998), implement a pre-existing fire spread model, propagating the fire front across a landscape. In particular, FARSITE is based on the BEHAVE system described earlier, and uses Huygen's principle (a vector-based approach) to model fire growth (Finney 1998). Mathematical analogue models use concepts from mathematics which exhibit similar behavior to draw comparisons to wildland fire spread. Common mathematical analogues include percolation methods, as discussed by Beer and Enting (1990), and cellular automata, first explored in this context by Albinet et al. (1986), which both use a raster-based approach to represent fire front propagation as a series of ignition events on a discrete lattice.

As described above, a common approach for real-time predictions of wildland fire spread dynamics is the use of empirical models, based on historical data and laboratory measurements (Sullivan 2009b). One possible alternative approach is machine learning (ML). The use of ML in wildfire modeling has seen a significant increase in popularity in recent years, with ML techniques used across tasks such as fuel characterization, risk assessment, fire behavior modeling, and fire management (Jain *et al.* 2020). ML has been used to predict fire severity metrics such as the burned area based on field conditions (Cortez and Morais 2007; Jain *et al.* 2020). Kozik et al. (2013) simulated fire spread using ML in conjunction with assimilating Geographic Information System data in order to reduce a priori uncertainties.

Here, we seek to use ML alone to explicitly model wildland fire spread dynamics based solely on data. In effect, ML in this context represents the development of an empirical model, but this development is driven by the ML algorithm rather than statistical analysis performed manually by researchers. Approaches that use ML alone to explicitly model wildfire dynamics are not common. Hodges and Lattimer (2019) used a convolutional neural network (CNN) to predict burn maps on simulated data evolving over six-hour intervals. Radke et al. (2019) used a CNN to predict how a fire front evolves over daily increments, but instead of using simulated data, they applied a CNN model to a single historical fire event.

The dynamics and morphology of a fire front can change significantly over time scales as short as minutes (Taylor 2020). So, instead of making a single prediction over a relatively large interval, the objective of this work is to model these short-term dynamics using ML techniques. To model the spatiotemporal dynamics of the fire front over an extended duration, we use an autoregressive process where one short-term prediction is used to generate the input for a subsequent prediction, and this process is repeated until as many time steps have taken place as desired. Autoregressive predictions are difficult to make reliably, as errors in the early predictions compound in subsequent predictions. We investigate how well commonly employed CNN models perform at this task. We investigate both atemporal CNNs, as well as temporally sensitive neural networks such as the Convolutional Long Short-Term Memory network (ConvLSTM) (Shi *et al.* 2015). This model combines the benefits of using convolutions with the benefits of tracking temporal relationships over time. For a few examples of ConvLSTMs being used in the geosciences, see (Arun *et al.* 2018; Zhong *et al.* 2019; Cruz and Bernardino 2019; Ienco *et al.* 2019; Zhai *et al.* 2020).

The reference data for the ML model was generated by an enriched mathematical analogue model. In contrast to observed wildland fire data, which is often satellite-based, simulated data provides access to spatio-temporally resolved data, and allows for the generation of arbitrarily large datasets. Further, the characteristics of the data can be tuned to test the performance of the model in response to specific features. Additionally, while the use of data from large-scale experimental fires was considered, the deep learning technique applied here requires a large body of data, containing hundreds to thousands of individual fires (Clements *et al.* 2019).

We present results on three datasets with increasing levels of complexity. The first dataset is simple, with only a single source of constant wind as a possible confounding factor, whereas the most complex dataset includes dynamic wind, complex elevation topologies, heterogeneous fuel distributions and spatially varying moisture content. While our data has qualities that represent realistic wildland fire scenarios, the analogue model employed in this work exhibits deficiencies in capturing all relevant physical processes at a quantitative level. Thus, the primary goal of this study is to demonstrate that current ML models can accurately represent the complex relationships that do exist in these wildland fire models. The proposed method is general and applicable to simulated datasets with even more complex dynamics, simulated datasets from physics-based models, or even experimental data.

## 2  Data

A mathematical analogue model is used to generate training data for the ML models. Details on the model are provided as Supplemental Material. The data consist of a 2D lattice grid of size (100, 100) cells, where each cell tracks the burn state and geographic data (such as density, terrain altitude, and moisture content) at that location. As the fire advances over time, a sequence of the burn state fields is produced. Each simulation is run until the fire is fully exhausted.

Three data sets were generated, each containing 1000 independent fire sequences. In each sequence, the environmental conditions of the field were randomly varied, with different datasets exhibiting different characteristics as seen in Table 1.

Table 1: Distinguishing characteristics of data sets used for training ML models.

| ID | Fuel Placement | Density | Moisture | Terrain | Wind |
|---|---|---|---|---|---|
| *wind* | Uniformly sparse | Heterogeneous | None | Flat | Fixed |
| *wind-slope* | Uniformly sparse | Heterogeneous | None | Planar | Fixed |
| *complex* | Patchy | Heterogeneous | Yes | patchy | Dynamic |

In each dataset, 50% of the cells (uniformly sampled) were filled with fuel of randomly assigned, normally distributed density ($\mu = 1$, $\sigma = 0.25$). These cases assumed no moisture content. The *wind* dataset was designed to test the performance of the ML models with a single confounding factor: a wind blowing in a constant direction for the entire duration of the fire and across the entire forest patch. The two vector components of the wind were generated with uniform distributions in the range $[-7, 7]$. The *wind-slope* dataset was designed to test the ML models' performance with multiple, potentially competing confounding factors by adding a planar slope to the terrain. Each of the 1000 fire sequences was given both a random wind (generated in the same manner as before) and a terrain with a slope of random grade (uniform distribution, $\theta \in [0, \pi/4]$) and azimuth (uniform distribution, $\phi \in [0, 2\pi]$). The final dataset is the *complex* dataset and includes all complexity available in the fire spread model. It featured patchy fuel placement which is representative of real forests (although vegetation growth patterns are not correlated with terrain as would be expected in a real forest). This vegetation was generated by placing circular areas of vegetation of radius 8 lattice units until the field was 70% filled. These areas were then masked, randomly clearing 40% of the cells, and a random density was assigned to the filled cells (normal distribution, $\mu = 1$, $\sigma = 0.25$). It also featured random moisture content (folded normal distribution, $\mu = 0$, $\sigma = 0.25$) and complex terrain elevation generated using the diamond-square algorithm (Fournier *et al.* 1982) with a maximum height of 50 lattice units. Finally, it also featured a change in the direction and magnitude of the wind at time-step 30 in the simulation (uniform distribution, components $\vec{U} \in [-12, 12]$).

The burnout time for a single fire is generally between 75 and 200 time-steps. Since the analog model is not calibrated, as described in the supplementary material, the length of a single time step is not exactly specified, but roughly correspond to dynamics that are representative of a scale of 5 minutes. Each cell in a simulated field corresponds to roughly 10 meters. For each case, a spot fire of radius 3 lattice units was initiated at a random, uniformly distributed location within the middle 50% of the field.

Representative sample outputs from these datasets are depicted in Figure 1. The first sample, taken from the *wind* dataset, features uniformly sparse, heterogeneous vegetation, no moisture, flat terrain, and wind. It demonstrates the response of the fire front to the wind, advancing with a bias towards the south. Sample 2, from the *wind-slope* dataset, is like sample 1, except it additionally features a planar, sloped terrain. Despite the forcing from the southeasterly wind, the magnitude of the wind is small, and the fire front responds to the significant slope and burns uphill to the south-southeast. Finally, sample 3, from the *complex* dataset, features patchy, heterogeneous vegetation, heterogeneous moisture content, procedurally generated terrain, and temporally varying wind. The fire begins in a low valley and responds to the wind by advancing with a westward bias. Eventually, it reaches regions of steep slope in the south and east of the domain and burns rapidly up these slopes. This behavior is augmented by the wind, which has shifted at $t = 30$ to a northwesterly direction.

Each of the 1000 fire sequences in each of the three datasets are partitioned into 200 sequences for a testing dataset and 800 sequences for a training dataset. An additional 200 fire sequences were made for the *complex* dataset and were used as validation dataset for hyperparameter training.

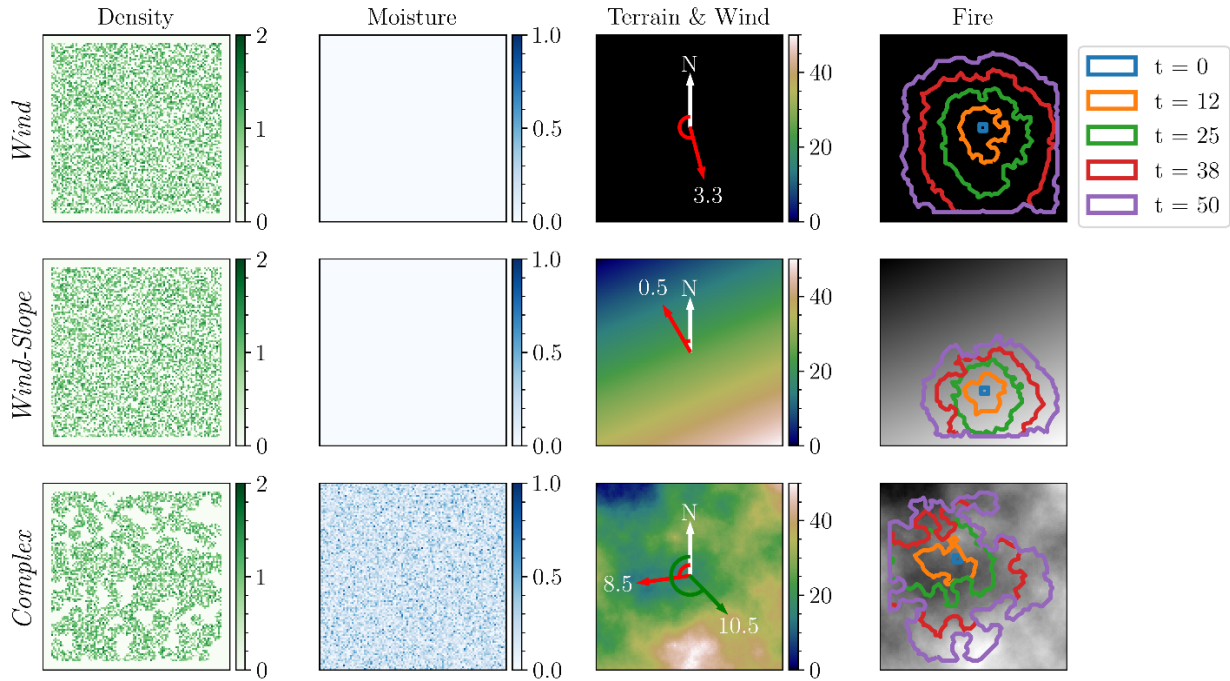

Figure 1: Representative sample input fields and fire spread output of the analogue model. Sample 1 is selected from the *wind* dataset, sample 2 is from the *wind-slope* dataset, and sample 3 is from the *complex* dataset; see Table 1 for details on the datasets.

## 3 Methods

Here, we present an overview of key aspects of the techniques used to train our machine learning models. For a detailed discussion of neural networks, see (Goodfellow *et al.* 2016).

The input to the model is a lattice of size (110, 110), where the inner (100, 100) region is filled with vegetation as described in Section 2, and the remaining border of 5 elements is set to zero. The DNN acts as a set of transformation on the input lattice, transforming it from a lattice specifying where the fire or burn scar currently is, into a lattice that specifies where the fire or scar will grow to at the next time step in the simulation. The most influential type of transformations applied in our DNNs are convolutions, in which a small lattice is convolved across the input lattice to create a transformed version of the input. For those unfamiliar with convolutions in this context, we recommend (Goodfellow *et al.* 2016).

The DNN is trained such that the transformations result in as accurate a prediction as possible, where the accuracy of the prediction is measured by the loss function. We use the mean squared loss function, which penalizes a model for every cell in the lattice it generates that does not match a corresponding cell in the lattice containing the ground truth.

## 3.1 Convolutional Neural Networks (CNNs)

A large number of CNNs used in practice follow a common pattern popularized by the autoencoder (Kramer 1991). An autoencoder is a DNN that consists of an encoder and a decoder (Figure 2a). The encoder attempts to reduce the lattice into a set of parameters that describe the variance seen in the input. The last layer in the encoder is called the bottleneck and is treated as input to the next stage: the decoder. The decoder takes the features in the bottleneck and regenerates the original lattice.

Instead of regenerating the input lattice, it is also possible to use the same network structure to generate a new lattice. In our case, that new lattice will be the location of the fire front in the next time step of the simulation. Thus, our CNN is a so-called autoencoder-like network. That is, it has the same structure as an autoencoder, but the predictive task is different. We use the autoencoder as a baseline that roughly corresponds to the work of Hodges and Lattimer (2019). The structure of our autoencoder-like model is shown in Figure 2a. It starts off with the input lattices of shape ($N_H = 110$, $N_W = 110$, $N_c$) with $N_c$ being the number of channels in the input lattice. Each channel provides a single value for each location in the lattice.

All three datasets have a *vegetation channel* specifying the location of unburnt fuel, a *scar channel* specifying the location of burnt fuel, a *front channel* specifying the location of the fire front at the previous time step and then two additional channels specifying horizontal and vertical wind velocities. The *wind-slope* and *complex* datasets includes a sixth channel specifying terrain height, and finally, the *complex* dataset includes a seventh channel specifying the initial moisture content of the vegetation.

All channels were normalized prior to training. The *wind channels* are the only channels that include negative values, and thus were normalized to have unit variance and zero mean. All other channels were normalized to range between 0.0 and 1.0 based on the min and max values seen throughout a single dataset.

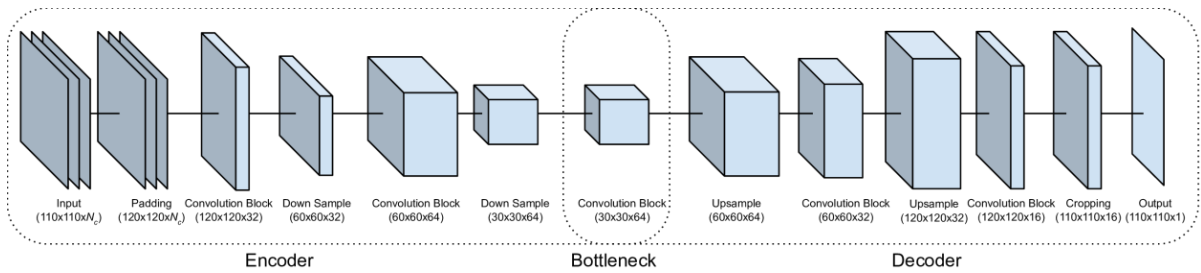

(a) Autoencoder-like CNN with two downsampling stages.

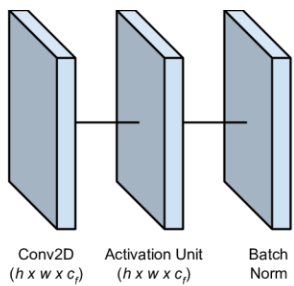

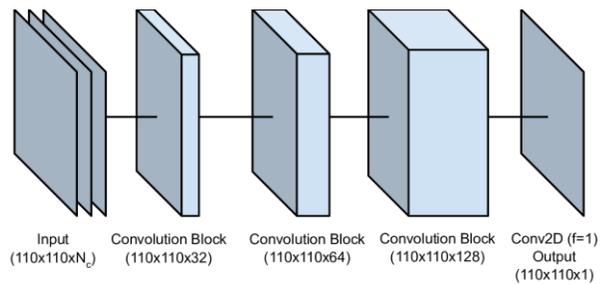

(b) Convolutional block.

(c) Simplified CNN.

Figure 2: CNN architectures employed in the present work. (a) Autoencoder-like CNN with two downsampling stages. (b) Convolutional block. $c_f$ is the number of filters applied in the convolution. (c) Simplified CNN.

The first stage of the autoencoder-like model takes the input lattices and transforms them into lattices of size (120, 120, $N_c$) by adding a zero-padding of 5 pixels around the border (this helps ensure the upsample operation described below does have aliasing issues). The model then applies its first convolution, followed by an activation unit, which ensures learned parameters are within acceptable bounds. A batch normalization operation is then added, which ensures the learned parameters have roughly the same range. Every time a convolution is applied in the model, it is always followed by these two additional steps, so all three operations are referred to as a *convolutional block* (Figure 2b). After the initial convolutional block, a downsampling operation is added that halves the $N_W$ and $N_H$ dimensions. After two steps of downsampling, the encoder ends with a final convolutional block of shape (30, 30, 64) that represents the bottleneck in the encoder/decoder structure. The decoder takes the output of the bottleneck and repeatedly applies an upsample operation, interpolating newly added values via a nearest-neighbors algorithm, before adding another convolutional block. The network is terminated via a single 2D convolution with a single filter, resulting in an output lattice of size (110, 110, 1). This lattice is the prediction made by the model.

We also investigate a simplified version of this CNN where we remove the downsampling operations, which results in the much simpler overall structure as shown in Figure 2c. The models are referred to as *CNN-Autoencoder* and *CNN-Simplified*, respectively. We performed a thorough hyperparameter search on both these models. See supplemental material. To correct for small errors that a CNN model may introduce over repeated predictions, we also create a *CNN-Thresholded* model which is the same as *CNN-Simplified* except it clips all values in the output lattice less than 0.025 to 0.0.

## 3.2 Convolutional Long Short-Term Memory (ConvLSTM) Network

The main weakness of the CNN approach is the lack of explicitly modeling the transient dynamics in the wildfire data. Simply put, the CNN is ignoring time. One of the most common deep learning methods for overcoming this limitation is to employ an RNN (Rumelhart *et al.* 1986) and in particular the LSTM (Hochreiter and Schmidhuber 1997). While successful in many domains, the LSTM is a poor choice for modeling lattices. An LSTM will attempt to identify a quadratic number of relationships between all elements in a lattice, which is typically not feasible. We overcome this by using a Convolutional LSTM (Shi *et al.* 2015), which only tries to identify how neighboring elements in the lattice are related.

The model is constructed as a sequential series of ConvLSTM blocks, and each ConvLSTM block takes as input a sequence of lattices, where each subsequent lattice in the sequence provides details about subsequent steps in time. The ConvLSTM model transforms the input sequence into a different representation by applying convolutions to the lattices. The output of the ConvLSTM block shares the same shape as the input, except the number of channels is set to the number of convolutions applied in the block. For example, the input to the first block is the input lattice of size (10, 110, 110, $N_c$).

The first convolution transforms that lattice into a new lattice of size (10, 110, 110, 20). That is, it replaces the $N_C$ values at each lattice location with 20 new values (20 was found to be optimal in our hyperparameter search). Subsequent blocks keep applying their transformations, until a final 3D convolution removes the time dimension and transforms the output of the last ConvLSTM block into the desired output with size (110, 110, 1). As with the CNN, we performed an extensive hyperparameter search described in the Supplemental Material.

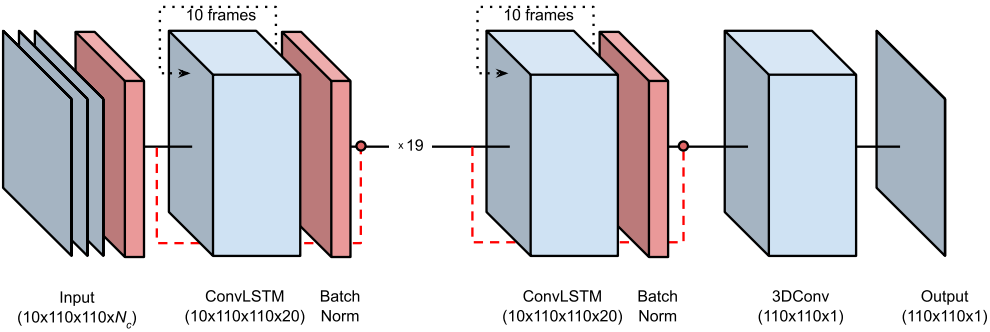

Figure 3: Structure of the ConvLSTM model. The black dotted line on the ConvLSTM blocks indicate the number of frames processed internally in the block, where $N_c$ indicates the number of channels in the input. The red dotted line corresponds to a skip link, which is implemented as a Conv3D layer with kernel size (1, 1, 1). Both the skip link and the batch normalization blocks are highlighted in red to indicate that they were not actually included in our best performing model but are common additions to ConvLSTMs that should be empirically evaluated.

### 3.3 Processing Consecutive Predictions and Post Processing

To measure how well our models estimate the dynamics of wildfire propagation, we measure how well the model performs over 50 consecutive autoregressive predictions. We start with input data as defined in Section 2. Given a trained model and an input lattice, we obtain the lattice that is the prediction the model transformed the input into. We use that prediction to update the training instance for the next autoregressive prediction via Algorithm 1.

---

**Algorithm 1** Autoregressive process for generating consecutive predictions

$datapoint_0 \leftarrow$ the initial training instance described in Section 2.
$prediction_0 \leftarrow$ the prediction for $datapoint_0$.
**for** $i$ in range(1:50) **do**
    $datapoint_i \leftarrow datapoint_{i-1}$.
    Subtract $prediction_{i-1}$ from the *vegetation channel* in $datapoint_i$.
    Replace any negative values in the new *vegetation channel* with 0.0.
    Add $prediction_{i-1}$ to the *scar channel* in $datapoint_i$.
    Set the *front channel* in $datapoint_i$ to be equal to $prediction_{i-1}$.
    Set *wind channels* in $datapoint_i$ to be equal to the wind at time step $i$.
    Remove the first time step from $datapoint_i$ (so it stays length 10).
    $prediction_i \leftarrow$ the prediction for $datapoint_i$.
**end for**

Preparing the consecutive input for the CNN is basically the same, except the training instances for the CNN only contain a single time step, so that the $i^{\text{th}}$ training instance does not contain any of the frames from the previous data instances.

### 3.4 Training Implementation

All models were built in TENSORFLOW 2.0 (Abadi *et al.* 2015). The Keras APIs (Chollet 2015) were used to build the model and the datasets. Models were trained on a single machine with 16 CPU cores and eight NVIDIA P100 GPUs. Keras's distributed strategy, MIRROREDSTRATEGY was used to coordinate multi-GPU training. The ADAM optimizer (Kingma and Ba 2015) was used. Training was done over 400 epochs, at which point all loss functions appear to have converged, though we acknowledge that detecting convergence is nontrivial and it is always possible that additional training epochs could potentially improve any of the models we trained.

### 3.5 Performance Metrics

For each cell in an input lattice, each of the models returns a floating-point value indicating how much vegetation in that cell will burn away at that point in time. One of the most common methods for measuring the quality of such a prediction is to compute the *mean squared error* (MSE) between the label and the predicted value. Given a prediction $\hat{\mathcal{Y}}$ and the corresponding ground truth, $\mathcal{Y}$:

$$MSE(\mathcal{Y}, \hat{\mathcal{Y}}) = \frac{1}{N_H N_W} \sum_{i=1}^{N_H} \sum_{j=1}^{N_W} |\hat{\mathcal{Y}}_{ij} - \mathcal{Y}_{ij}|^2 \qquad (1)$$

Note that the MSE metric is quite demanding and is used as our *loss function* during training. For example, a model might do a good job predicting the general shape of the fire front but might make errors in how quickly that front spreads. The MSE will be high simply because the fire front was in the wrong location and the fact that it was generally the correct shape and size is ignored by the metric. Thus, we also measure the total amount of fire at each moment in time, the summed total error, $STE$:

$$\text{STE}(\mathcal{Y}, \hat{\mathcal{Y}}) = \left| \sum_{i=1}^{N_H} \sum_{j=1}^{N_W} \hat{\mathcal{Y}}_{ij} - \sum_{i=1}^{N_H} \sum_{j=1}^{N_W} \mathcal{Y}_{ij} \right|. \qquad (2)$$

The STE metric disregards the exact shape of the prediction and focuses on the overall quantity of the prediction. So, the combination of MSE and STE provides insight into how well the models predict the correct location of the fire as well as the correct size of the fire.

Both MSE and STE are regression-based statistics. We also include metrics that focus on binary classification to determine where fire is or is not (without regard for how much fire is there). This is done by introducing a threshold, below which a cell is said to be not on fire and above which a cell is said to be on fire. We use a threshold of 0.1, which captures most cells legitimately on fire.

We use the Jaccard Similarity Coefficient (JSC) (Filippi *et al.* 2014) to measure the quality of the classifications. JSC grows as the intersection between the prediction and the ground truth grows, resulting in a maximal score when the intersection and union are identical. It can be defined straight-forwardly with indicator variables. Let $I^\theta$ be an indicator variable that is 1.0 when $\theta$, a predicted value or a ground truth value, is greater than the 0.1 classification threshold. JSC is then defined as:

$$JSC(\mathcal{Y}, \hat{\mathcal{Y}}) = \frac{\cap (\mathcal{Y}, \hat{\mathcal{Y}})}{\cup (\mathcal{Y}, \hat{\mathcal{Y}})} \qquad (3)$$

$$\cap\left(\mathcal{Y},\hat{\mathcal{Y}}\right) = \sum_{i=1}^{N_H}\sum_{j=1}^{N_W} I^{\mathcal{Y}_{ij}} \wedge I^{\hat{\mathcal{Y}}_{ij}} \tag{4}$$

$$\cup\left(\mathcal{Y},\hat{\mathcal{Y}}\right) = \sum_{i=1}^{N_H}\sum_{j=1}^{N_W} I^{\mathcal{Y}_{ij}} \vee I^{\hat{\mathcal{Y}}_{ij}} \tag{5}$$

JSC ranges between 0.0, in which the prediction and the ground truth do not overlap at all, and 1.0 in which the ground truth and prediction are perfectly aligned. A weakness of the JSC metric is that it is not sensitive to how far off a prediction is in space. For example, the JSC score will be the same for a misclassification that placed a fire front one pixel away as it would if the front was 10 pixels away.

The *Shape Agreement* metric (SA) helps resolves this by explicitly considering how far apart an errant prediction was from the ground truth in time. A fire front errantly placed one cell away from the correct location is penalized less than an error placing the front 10 cells away (provided that the correct location is eventually burned). We refer the reader to (Filippi *et al.* 2014) for additional details on the mathematical description of this metric.

We apply each of these metrics to two different predictions. First, the prediction of where the fire-front is at each time step. Second, the prediction of where the fire-scar is at each time step. We refer to these as *front* or *scar* results, respectively.

### 3.6 Bootstrapping Confidence Intervals

When determining whether one model performs better than another model on a given metric, it is important to determine whether the difference in the metric is due to the random chance of a sampling bias in the data. To account for this, we use bootstrapping (Efron and Tibshirani 1993).

For each testing dataset across the three experimental conditions (*wind*, *wind-slope*, *complex)*, we create 20 *bootstrapping* datasets. Each bootstrapping dataset is constructed by randomly sampling fire sequences from the 200 samples in each testing dataset, with replacement, until the bootstrapping dataset contains the same number of fire sequences as the testing dataset. Once the sample is complete, all the data points normally generated from the sampled fire sequences are collated into a single bootstrapping dataset.

All metrics (MSE, STE, JSC, SA) are computed for each of the 20 bootstrapping datasets, giving a distribution of the metric across the bootstrapping samples. 50% confidence intervals are computed by taking the 0.25 and 0.75 quantiles from each distribution, and 90% confidence intervals are computed by taking the 0.05 and 0.95 quantiles (20 was chosen as that is the smallest sample size that has a distinct 0.95 quantile).

## 4 Results

### 4.1 CNN Results

As mentioned in Section 3.1, we investigated three variants of non-temporal CNN models: the *CNN-Autoencoder* model, the *CNN-Simplified* model and the *CNN-Thresholded* model. Due to space constraints, we summarize the major results of these three models. The *CNN-Autoencoder* model performed by far the worst on classification, having a Jaccard Similarity Coefficient (JSC) score dramatically lower than the other two models. For example, on the most difficult dataset (*wind-slope*), the *CNN-Autoencoder* model had a JSC of approximately 0.8 on the initial prediction, whereas the other models had JSC scores of approximately 0.99 on the initial prediction. While the quantitative values change across the other datasets and metrics, qualitatively the results are the same: *CNN-Autoencoder* performs poorly. *CNN-Simplified* performed similarly to *CNN-Thresholded* on the *wind-slope* and *complex* datasets but was significantly outperformed on the *wind* dataset. Thus, overall, the *CNN-Thresholded* model performed the best. In the next section, we provide a detailed comparison of its performance with the best performing ConvLSTM model.

## 4.2  ConvLSTM Results

Figure 4 provides an example for a single prediction of a single fire sequence in the *complex* dataset. The *ConvLSTM* model is used to make this prediction, but there are similarly behaving examples for the CNN models as well. There are four rows of results, corresponding to predictions at time instances $t = \{0, 14, 29, 49\}$. Notice that the fire front is not a single contiguous front, but instead has a complicated and disjoint structure.

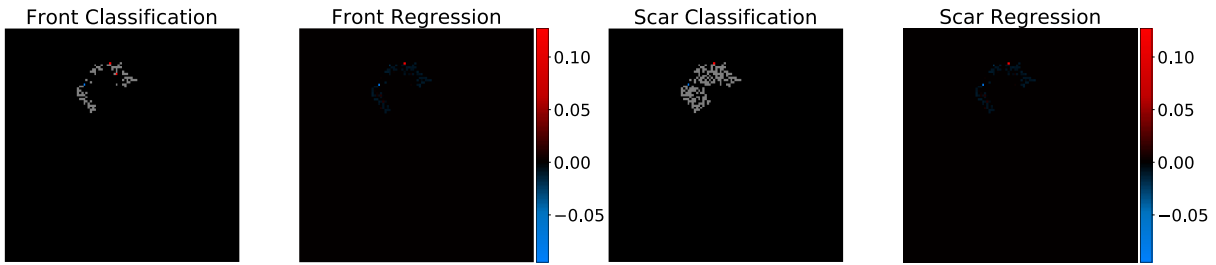

(a)  Prediction at $t = 0$: Very few errors at initial instance.

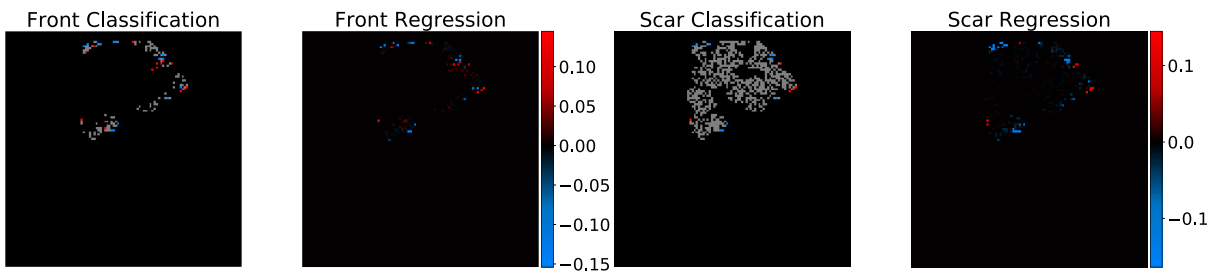

(b)  Prediction at $t = 14$: Errors starting to accumulate.

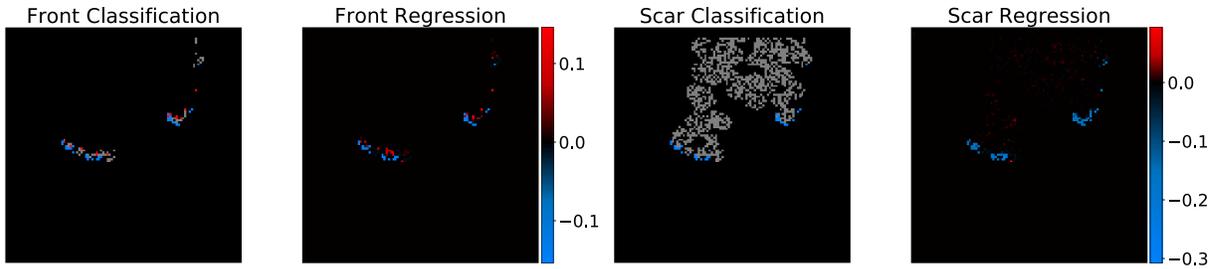

(c) Prediction at $t = 29$: Indication of overprediction of propagation speed of fire front.

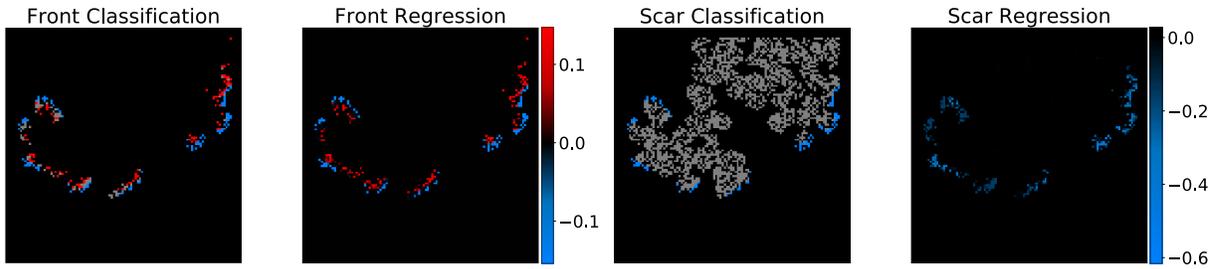

(a) Prediction at $t = 49$

Figure 4: Visualization of four predictions from a single predicted fire sequence. *Classification* graphs show the classification errors. Red cells indicate a prediction of fire when there was no fire, and blue cells indicate a prediction of no fire when there was fire. *Regression* graphs show difference between the amount of fire predicted and the actual amount of fire.

The first prediction at $t = 0$ is nearly perfect, though the *Front Classification* shows there is a single location colored in red, indicating that the predicted value was too large, and a single location colored in blue, indicating that the predicted value was too small. The grey and black cells correspond to correct predictions fire and no fire, respectively.

For the fire-spread prediction at $t = 14$, there are now more errors, as is expected, but the model provides reasonable predictions of the fire-spread dynamics. For the consecutive prediction at $t = 29$, there is clear evidence that the model has made errors in the speed at which it propagates the fire front. The red cells show where the model errantly predicts the fire location, whereas the blue cells show the actual fire location.

We can also visualize the performance of the ConvLSTM model via a comparison of the predicted and ground truth fire fronts. Figure 5 depicts the growth of these fronts for one sample case from each dataset.

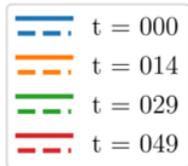

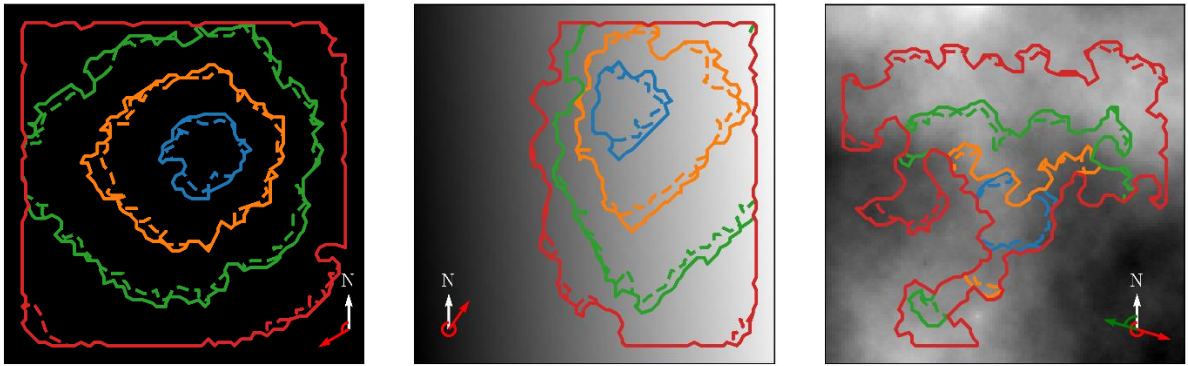

(a) *Wind* sample  (b) *Wind-slope* sample  (c) *Complex* sample

Figure 5: Visualizations of fire front growth for sample cases from each dataset. The solid line represents the ground truth fire front, while the dashed line represents the prediction of the ConvLSTM model. Fronts are overlaid on a heatmap of the terrain altitude. Wind direction is indicated by the red arrow. In the case of the *complex* sample, the red arrow indicates the wind direction for $0 \leq t < 30$, while the green arrow indicates the wind direction for $t \geq 30$.

In these plots, the fire front is determined by computing an alpha shape (Edelsbrunner *et al.* 1983) of the cells which are either burning or burnt, with $\alpha = 0.5$. It is clear from these plots that the ConvLSTM model is accurately predicting both the shape and speed of the fire front from a qualitative standpoint. It appears to respond well even across highly varying environmental conditions.

### 4.3 Quantitative analysis

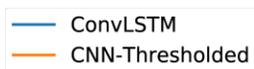

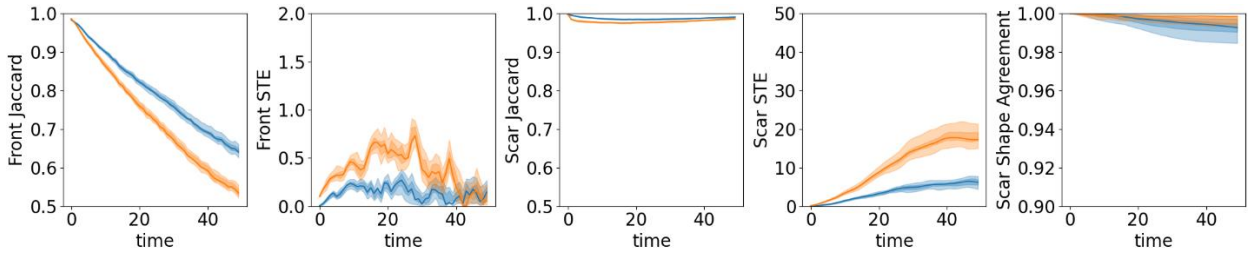

(a) *Wind*

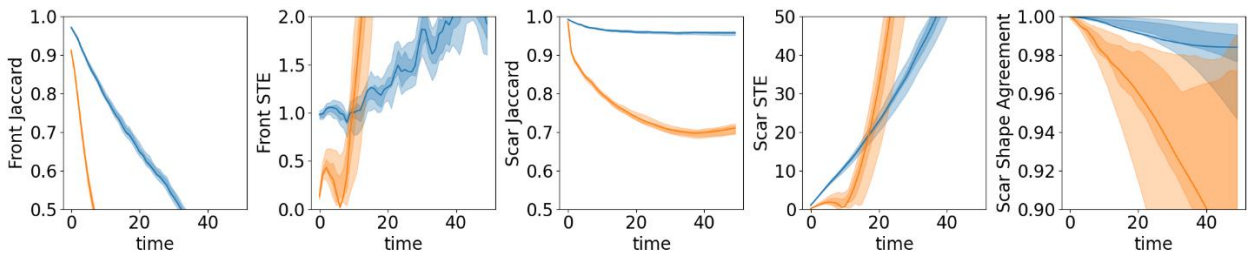

*(b) Wind-Slope*

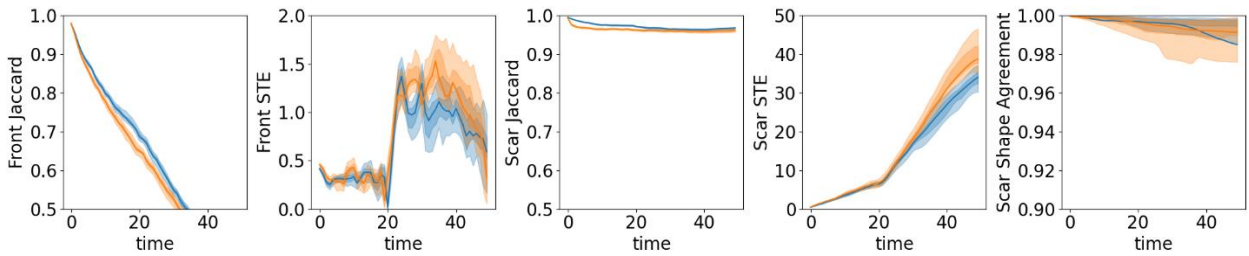

(c) *Complex*

Figure 6: Comparison between the best performing ConvLSTM and the best performing CNN. The horizontal axis on all graphs corresponds to the 50 total consecutive predictions. Shaded regions correspond to 50% and 90% confidence intervals.

Figure 6 compares the results of the ConvLSTM model with *CNN-Thresholded*. When measured by the JSC, the classification performance of the ConvLSTM model clearly dominated the performance of the CNN on the *wind* and *wind-slope* datasets, and maintained an advantage on the *complex* dataset, though by a smaller degree. This advantage was seen both when the entire fire scar was being predicted, as well as when just the fire front was being predicted.

Note that predicting the fire front is a more challenging task than predicting the fire scar, in part because once a lattice element is in the scar, it will always be in the scar. The SA metric helps by penalizing a prediction proportionally to how far off in space that a location was said to be on fire and showed that the ConvLSTM outperformed the CNN, but in general, the metric was too sensitive to sampling bias to make definitive conclusions on the other two datasets.

The STE score provides details of how well each model predicted the overall size of the fire front and the scar. Again, the results demonstrate that the ConvLSTM outperforms the CNN, consistently having lower STE errors than the CNN. This indicates that the ConvLSTM is not only better at predicting the locations and times that fires occur, but also the overall amount of fire. The main notable exception is on the *wind-slope* dataset. For the first 10 predictions, the CNN has significantly less STE error than the ConvLSTM. We suspect that the CNN is attempting to correctly estimate overall fire size at the expense of finding the exact location of fire spread, so it performs well early on, but then accumulated errors degrade future performance.

## 4.4 Unexpected Results

One of the most surprising results was the fact that the *wind-slope* dataset was the hardest dataset to classify. This was the dataset that likely had the least coverage in the training corpus in terms of interacting forces on the fire front. In the *wind* dataset, there are approximately two training samples for each of the 360 possible degrees of wind direction (on average), meaning that most validation samples likely had a training sample with a similar wind direction (though the wind velocity could still be quite different). However, with the *wind-slope* dataset, the joint space of wind direction and slope direction has far less coverage, so a validation example with a given wind and terrain slope likely has no exact counterparts in the training data. The *complex* dataset does not suffer from this issue as much since its slopes are not planar. The more realistic elevation maps it contains have good coverage in the training data. Further, the elevation and patch vegetation act as constraints to where the fire can realistically spread, further simplifying classification. As such, while the *complex* dataset is the most realistic, it is not the most difficult to classify.

The ConvLSTM model outperformed the CNN model most significantly on *wind-slope*. This suggests that as the difficulty of the underlying prediction increases, the benefit of the ConvLSTM Increases. We hypothesize that the ConvLSTM acts as a type of error correction, in which errant predictions at one consecutive prediction are ignored in subsequent consecutive predictions by the memory cells of the LSTM. This is consistent with the observation that the CNN model generally had more variance on the bootstrapping datasets than the ConvLSTM, as the error correction would allow the ConvLSTM model to perform more consistently across the bootstrapping samples. Recall that the *CNN-Simplified* model was improved to be the *CNN-Thresholded* model by turning small predictions into zeros—no such improvement was required for the ConvLSTM model to perform well.

It was also surprising that simpler variants of both the ConvLSTM and the CNNs outperformed more complicated versions. For example, in the ConvLSTM, we found that residual links and batch normalization layers reduced overall performance. For the CNNs, we found that the complex autoencoder-like model (including downsampling and upsampling layers) performed worse than a simpler CNN with just a few stacked convolutional layers.

# 5   Conclusions and Future Work

The primary contribution made in this work is demonstrating that the Convolutional Long Short-Term Memory deep neural network can capture the transient dynamics of wildland fires as described by a class of widely used numeric models. To the best of our knowledge, this work represents one of the first attempts to model such data, along with the previous work of Hodges and Lattimer (2019) and Radke et al. (2019). Previous work focused on making a single prediction for a relatively large time step (hours or days). We instead focus on an autoregressive process, making predictions mere minutes into the future, and consecutively repeating those prediction to extend to larger periods of time.

As sufficiently detailed empirical wildland fire data does not currently exist, we demonstrate the efficacy of the ConvLSTM model on several corpuses of simulated data generated by an analogue model. Three distinct data corpuses were tested, each with an increasing amount of complexity. We compare the performance of the ConvLSTM to convolutional neural networks. After performing a hyperparameter search to empirically identify the best-performing ConvLSTM and CNN, we demonstrate that the ConvLSTM provides improved predictions of the fire-front dynamics.

Overall, these results are an encouraging signal for using deep neural networks to model wildland fire propagation, though there is plenty of future work that remains before Machine Learning models can be employed in an operational setting. Currently, the types of observations required to drive the models proposed herein do not exist. Further work involves demonstrating that these techniques can be employed when training data is sparse, both temporally and spatially. Training on real-world datasets would allow the learned model to provide real-world physical insights about fire propagation in a way that a model trained on simulated data cannot. Finally, the predictive performance of the underlying DNN models could likely be improved. Our work demonstrated that relatively simple models are efficacious, but there are a large number of possible modifications and different approaches that could be investigated, for example: replacing the loss function with a probabilistic model (Dillon *et al.* 2017); using attention based mechanisms (Vaswani *et al.* 2017; Ramachandran *et al.* 2019; Cordonnier *et al.* 2019); or investigating other popular networks like the CNN-LSTM (Donahue *et al.* 2015; Sønderby *et al.* 2020).

## Acknowledgements

We would like to thank the members of Google's AI For Weather Team: Jason Hickey, Cenk Gazen, Shreya Agrawal, as well Qing Wang, Yi-Fan Chen, John Anderson, Carla Bromberg, Stephan Hoyer and Casper Sonderby for their guidance which accelerated the modeling and experimentation, and Martin Mladenov for discussions on extending our work in the context of TENSORFLOW Probability.

## Supplementary Resources

The TENSORFLOW code for the mathematical analogue model is available at https://github.com/IhmeGroup/Wildfire-TPU and can be used for generation of data. All data used for training, testing, and validation the ML-models are available through the Kaggle-data repository at https://www.kaggle.com/johnburge. Sample codes for the generation of the CNNs and ConvLSTM used in this work are available upon request to the author.

## Statement of Conflicting Interest

This research did not receive any specific funding. The authors declare there are no conflicts of interest.

# Supplementary Material For:

# Convolutional LSTM Neural Networks for Modeling Wildland Fire Dynamics

John Burge[a,c], Matthew Bonanni[b], Matthias Ihme[a,b], and R. Lily Hu[a]

[a]Google Research, Mountain View, CA 94043, USA
[b]Department of Mechanical Engineering, Stanford University, Stanford, CA 94305, USA
[c]Corresponding Author. Email: lawnguy@google.com

## 6  Analogue Model

1    The reference data for the ML model was generated with a wildfire spread model. This
2    mathematical analogue model uses a percolation method at its core (Beer and Enting 1990), but
3    additionally takes into consideration effects of fuel density, moisture content, slope, and wind,
4    informed by insights from popular quasi-empirical models. The model is implemented in
5    TENSORFLOW (Abadi *et al.* 2015), taking advantage of the programming-specific data structures
6    and linear algebra, and enabling the simultaneous calculation of large ensembles. By
7    implementing this model in TENSORFLOW, we also allow for more seamless integration of the data
8    generation and ML modules as many ML models are also implemented in TENSORFLOW (or
9    similar infrastructures).

10   In the model, the 2D domain with width $W$ and height $H$ is discretized using a lattice with
11   equidistant spacing $\Delta$. This enables the field variables to be represented by matrices of dimension
12   $N_W \times N_H$, where $N_W = W/\Delta$, $N_H = H/\Delta$. At each discrete time step $t + \Delta t$, each lattice cell
13   absorbs heat from its neighbors, and the amount of heat is weighted by the terrain and wind conditions at those neighbors. Denoting the time-dependent heat content of lattice cell $(i,j)$ by $Q_{i,j}(t)$, this model can be expressed by the following discrete form:

$$Q_{i,j}(t + \Delta t) = Q_{i,j}(t) + \sum_{k=-N_R}^{N_R} \sum_{l=-N_R}^{N_R} \Omega_{i,j,k,l}(t) K_{k,l} \quad (1)$$

where $N_R$ represents the discrete number of neighbors that directly influence the local cell $(i,j)$. The region of influence depends on the size of the cell's neighborhood (Beer and Enting 1990). $\Omega_{i,j,k,l}(t)$ is the time-dependent heat accumulation rate, and $K_{k,l}$ is a Boolean matrix representing the interacting neighborhood.

Equation 1 is effectively a discrete convolution of the spatio-temporally varying heat-accumulation kernel $\Omega$ over the image matrix $K$. $\Omega$ is evaluated taking into consideration the terrain, wind, and ignition state (Beer and Enting 1990):

$$\Omega_{i,j,k,l} = \Phi_{i,j,k,l} \odot \Psi_{i,j,k,l} \odot L_{i,j} \quad (2)$$

where $\odot$ denotes the Hadamard (or element-wise) product. The slope factor $\Phi_{i,j,k,l}$, accounting for the local change in terrain, is given as:

$$\Phi_{i,j,k,l} = \exp\left(\alpha_s \frac{(k,l)^T}{\sqrt{k^2+l^2}} \cdot \vec{S}_{i+k,j+l}\right) \quad (3)$$

where $\alpha_s$ is the slope sensitivity factor, and $\vec{S}$ is the gradient of the terrain altitude $Z$

$$\vec{S}_{m,n} = \frac{1}{2\Delta}(Z_{m+1,n} - Z_{m-1,n}, Z_{m,n+1} - Z_{m,n-1})^T. \quad (4)$$

The wind factor, $\Psi_{i,j,k,l}$, is modeled to yield an elliptical fire front in a continuous, homogeneous field, as described by Catchpole et al. (Catchpole *et al.* 1982):

$$\Psi_{i,j,k,l} = 1 + \alpha_w \frac{ab^2\varepsilon \cos(\Theta_{k,l}) + ab\Gamma_{k,l}}{a^2 \sin^2(\Theta_{k,l}) + b^2 \cos^2(\Theta_{k,l})}, \tag{5}$$

with

$$\Gamma_{k,l} = \sqrt{a^2 \sin^2(\Theta_{k,l}) - a^2\varepsilon^2 \sin^2(\Theta_{k,l}) + b^2 \cos^2(\Theta_{k,l})}, \tag{6a}$$

$$\Theta_{k,l} = \cos^{-1}\left(\frac{\vec{U}}{\|\vec{U}\|} \cdot \frac{(k,l)^T}{\sqrt{k^2 + l^2}}\right), \tag{6b}$$

$$\zeta = 1 + \frac{1}{4}\|\vec{U}\|, \tag{6c}$$

$$\varepsilon = \sqrt{1 - \zeta^{-2}}, \tag{6d}$$

$$a = \frac{\|\vec{U}\|\Delta t}{1 + \varepsilon}, \tag{6e}$$

$$b = \frac{a}{\zeta}, \tag{6f}$$

where $\Theta_{k,l}$ is the angle between the wind direction and the vector towards the lattice cell, $\alpha_w$ is the wind sensitivity factor, $\zeta$ is the ellipse length-to-width ratio, $\varepsilon$ is the ellipse eccentricity, and $a$ and $b$ are the ellipse semi-minor and semi-major axes, respectively. While this formulation assumes a spatially uniform wind $\vec{U}$, it can be directly extended to a non-uniform wind $\vec{U}_{i,j}$, where $\zeta$, $\varepsilon$, $a$, and $b$ also become 2D tensors instead of scalars, and $\Theta_{k,l}$ and $\Gamma_{k,l}$ become 4D tensors $\Theta_{i,j,k,l}$ and $\Gamma_{i,j,k,l}$.

The ignition state tensor, $L_{i,j}$, is activated when sufficient heat has accumulated, and deactivated when the burn duration has been reached. This can be represented by a product of Heaviside functions:

49 $$L_{i,j}(t) = \mathfrak{H}(Q_{i,j}(t) - Q_{i,j}^{ign})\mathfrak{H}\left(D_{i,j} - (t - t_{i,j}^{ign})\right), \tag{7}$$

50 where $t_{i,j}^{ign}$ is the time of ignition and $D_{i,j}$ is the burn duration, which is calculated as:

51 $$D_{i,j} = D^0 R_{i,j}, \tag{8}$$

52 with $D^0$ being the nominal burn duration and $R_{i,j}$ the local fuel density. The ignition time $t_{i,j}^{ign}$ is

53 incremented over time in the unburnt cell until sufficient heat is accumulated for ignition to occur:

54 $$t_{i,j}^{ign}(t + \Delta t) = t_{i,j}^{ign}(t) + (\Delta t)\mathfrak{H}\left(Q_{i,j}^{ign} - Q_{i,j}(t)\right), \tag{9}$$

55 where $Q_{i,j}^{ign}$ is a tensor representing the amount of heat required to ignite the lattice cell. Following

56 Rothermel's formulation (Rothermel 1972), this quantity is evaluated as:

57 $$Q_{i,j}^{ign} = Q_{i,j}^{ign,d} + \alpha_m M_{i,j}, \tag{10}$$

58 where $Q_{i,j}^{ign,d} = Q^0 R_{i,j}$ is the heat required to ignite the cell once dried, $Q^0$ is the nominal ignition

59 heat, $R_{i,j}$ is the density, $\alpha_m$ is the moisture sensitivity factor, and $M_{i,j}$ is the moisture content. Note

60 that $L$ and $t^{ign}$ both have a dependency on $Q$, which is resolved as $L$ is initialized with a given fire

61 state, $Q$ is initialized to $Q^{ign}$ at the affected cell, and $Q$, $L$, and $t^{ign}$ are then updated sequentially

62 as described.

63 The kernel $K_{k,l}$, appearing in Equation 111, represents the non-local interaction with the

64 surrounding neighborhood and is computed as

65 $$K_{k,l} = \mathfrak{H}\left(N_R - \sqrt{k^2 + l^2}\right) - \delta_{k0}\delta_{l0} \quad \text{for } k, l \in [-N_R, N_R] \tag{11}$$

66 where $\delta_{mn}$ is the Kronecker delta to exclude self-interaction.

67   This model has not been calibrated to reproduce real wildfire behavior; rather, it was
68   designed to emulate the behavior of fire spread. This simplified model yields data that is useful for
69   the training of ML models. Neural networks that can successfully model this data are expected to
70   be able to model calibrated data. To this end, values for the given parameters were selected to
71   target the percolation threshold, yielding fires that exhibit heterogeneous spread behavior without
72   burning out prematurely. The parameter values used for the simulations in this work are presented
73   in Table 1.

Table 2: Analogue model parameters used in this work.

| Name | Description | Value |
|---|---|---|
| $\Delta$ | Lattice spacing | 1 |
| $N_W$ | Number of points across field width | 110 |
| $N_H$ | Number of points across field height | 110 |
| $N_R$ | Neighborhood size | 3 |
| $D^0$ | Nominal burn duration | 3 |
| $Q^0$ | Nominal ignition heat | 3 |
| $\alpha_m$ | Moisture sensitivity | 1 |
| $\alpha_s$ | Slope sensitivity | 0.7 |
| $\alpha_w$ | Wind sensitivity | 2 |

74   A representative sample output from this model is presented in Figure 1. This simulation,
75   from the *complex* dataset described in the text, exercises all of the features of the analog model.
76   The flame front has a complex shape due to the heterogeneity of the fuel and terrain. At time step
77   $t = 20$, it can be seen that the front has responded to the westerly wind. At $t = 40$, the wind has
78   changed direction and the fire has responded by advancing in the southwest direction. We also
79   note that it has continued to burn to the northeast despite the adverse wind, because of the favorably
80   sloped terrain.

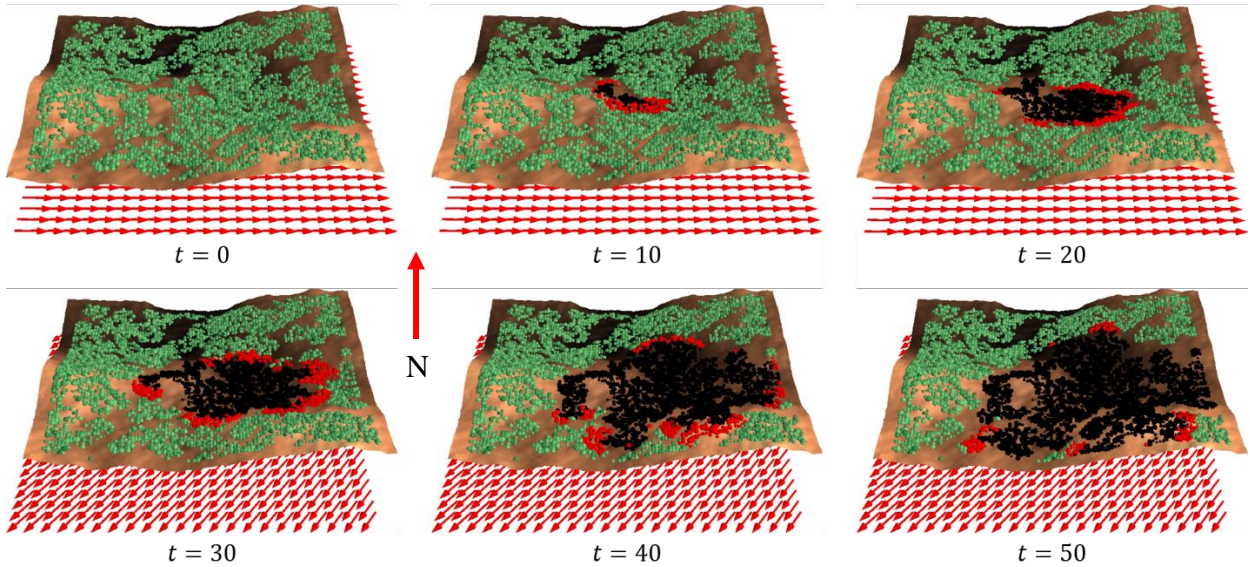

Figure 7: Visualization of a representative sample output from the *complex* dataset at successive time steps. The ground color corresponds to terrain altitude, and the arrows depict spatially constant but temporally changing wind direction.

## 7 Hyperparameters

One of the main challenges in using deep neural networks (DNNs) is the need to specify hyperparameters. These are parameters that help govern how a neural network is trained. The following are the most significant hyperparameters for the CNN models:

- `lr` $\in \{10^{-2}, 10^{-3}, 10^{-4}, 10^{-5}, 10^{-6}\}$ indicates the learning rate. This influences how quickly the training process iteratively makes updates to the model.
- `ks` $= \{3, 5\}$ specifies the kernel size for both the height and width dimensions of the convolutions.
- `au` $\in \{$`ReLU`, `LeakyReLU`$\}$ specifies which activation unit to use after convolutional blocks.
- `ds` $\in \{$`yes`, `no`$\}$ indicates whether to include downsampling stages.

- bn ∈ {yes, no} indicates whether batch normalization stages are included in the convolutional block.
- cb ∈ {1, 2, 3} specifies the number of convolutional blocks (and downsampling stages, if enabled) that are included.
- nf is a list of length cb specifying the number of filters in each of the convolutional blocks. For example, nf = [16, 32, 64] indicates that the three convolutional blocks in the encoder contain 16 filters, then 32 filters and finally 64 filters, whereas the convolution blocks in the decoder would contain 64 filters, then 32 filters, then 16 filters.
- bf ∈ {8, 16, 32, 64, 128, 256} specifies the number of filters in the bottleneck. We always set this to the number of filters in the last convolutional block.

Ideally, we would perform a hyperparameter search independently for each dataset, however, this is not feasible. So, to reduce the amount of training required, we performed the hyperparameter search on the *complex* dataset, and then used those same hyperparameters across all datasets. It is possible that separate hyperparameter searches for each dataset would result in stronger models.

Our final settings for these hyperparameters are: {ds = no, bn = no, cb = 3, nf = [32, 64, 128], bf = 128, ks = 5, au = ReLU}. In all, we needed to train approximately 30 different models with various hyperparameter settings to identify this representative case.

Similarly, we empirically identified the best hyperparameter for the ConvLSTM model:

- nb ∈ (0, 10] specifies the number of stacked blocks.
- fpb ∈ (0, 20] specifies the number of filters to use per ConvLSTM block.
- bn ∈ {yes, no} specifies whether batch normalization was used after each ConvLSTM block.

- `do` ∈ {0.0, 0.01, 0.05} specifies how much dropout to use in the convolutional layers (i.e., what fraction of network nodes should be randomly reset to 0.0 after each epoch).
- `sl` ∈ {yes, no} specifies whether skip links are added for each of the stacked blocks.

The final representative ConvLSTM model had the hyperparameter settings: {`nb` = 10, `fpb` = 20, `bn` = no, `do` = 0.0, `sl` = no}.